\documentclass[10pt,twocolumn,letterpaper]{article}

\usepackage{cvpr}
\usepackage{times}
\usepackage{epsfig}
\usepackage{graphicx}
\usepackage{amsmath}
\usepackage{amssymb}
\usepackage{paralist}

\usepackage{multirow}
\usepackage{color}
\usepackage[pagebackref=true,breaklinks=true,letterpaper=true,colorlinks,bookmarks=false]{hyperref}

\newcommand{\wqz}[1]{\textcolor{black}{#1}}
\newcommand{\abc}[1]{\textcolor{black}{#1}}
\newcommand{\abcn}[1]{\textcolor{black}{#1}}
\newcommand{\abcnn}[1]{\textcolor{black}{#1}}
\definecolor{purple}{rgb}{0.87, 0.0, 1.0}
\newcommand{\wqzp}[1]{\textcolor{black}{#1}}
\newcommand{\wqzpp}[1]{\textcolor{black}{#1}}

\newcommand{\reffig}[1]{Fig.~\ref{#1}}

\cvprfinalcopy % *** Uncomment this line for the final submission

\def\cvprPaperID{1855} % *** Enter the CVPR Paper ID here

\ifcvprfinal\fi
\begin{document}

%%%%%%%%% TITLE
\title{CNN$+$CNN: Convolutional Decoders for Image Captioning}

\author{Qingzhong Wang and Antoni B. Chan\\
Department of Computer Science,
City University of Hong Kong\\
{\tt\small qingzwang2-c@my.cityu.edu.hk, abchan@cityu.edu.hk}
% \and
% Antoni B. Chan\\
% Department of Computer Science\\
% City University of Hong Kong\\
% {\tt\small abchan@cityu.edu.hk}
}

\maketitle
%\thispagestyle{empty}

%%%%%%%%% ABSTRACT
\begin{abstract}
   Image captioning is a challenging task that combines the field of computer vision and natural language processing. A variety of approaches have been proposed to achieve the goal of automatically describing an image, and recurrent neural network (RNN) or long-short term memory (LSTM) based models dominate this field.  However, RNNs or LSTMs cannot be calculated in parallel and ignore the underlying \abc{hierarchical} structure of a sentence. In this paper, we propose a framework that only employs convolutional neural networks (CNNs) to generate captions. Owing to parallel computing, our basic model is around $3\times$ faster than NIC (an LSTM-based model) during training time, while also providing better results. We conduct extensive experiments on MSCOCO and investigate the influence of the model width and depth. %hyper-parameters. 
   Compared with LSTM-based models that apply similar attention mechanisms, our proposed models achieves comparable scores of BLEU-1,2,3,4 and METEOR, and higher scores of CIDEr. We also test our model on the paragraph annotation dataset \cite{hierlstm}, and get higher CIDEr score compared with hierarchical LSTMs. 
\end{abstract}

%%%%%%%%% BODY TEXT
\vspace{-0.2in}
\section{Introduction}\label{sec1}

\begin{figure}[t]
\centering
\includegraphics[width=0.4\textwidth, angle=-90]{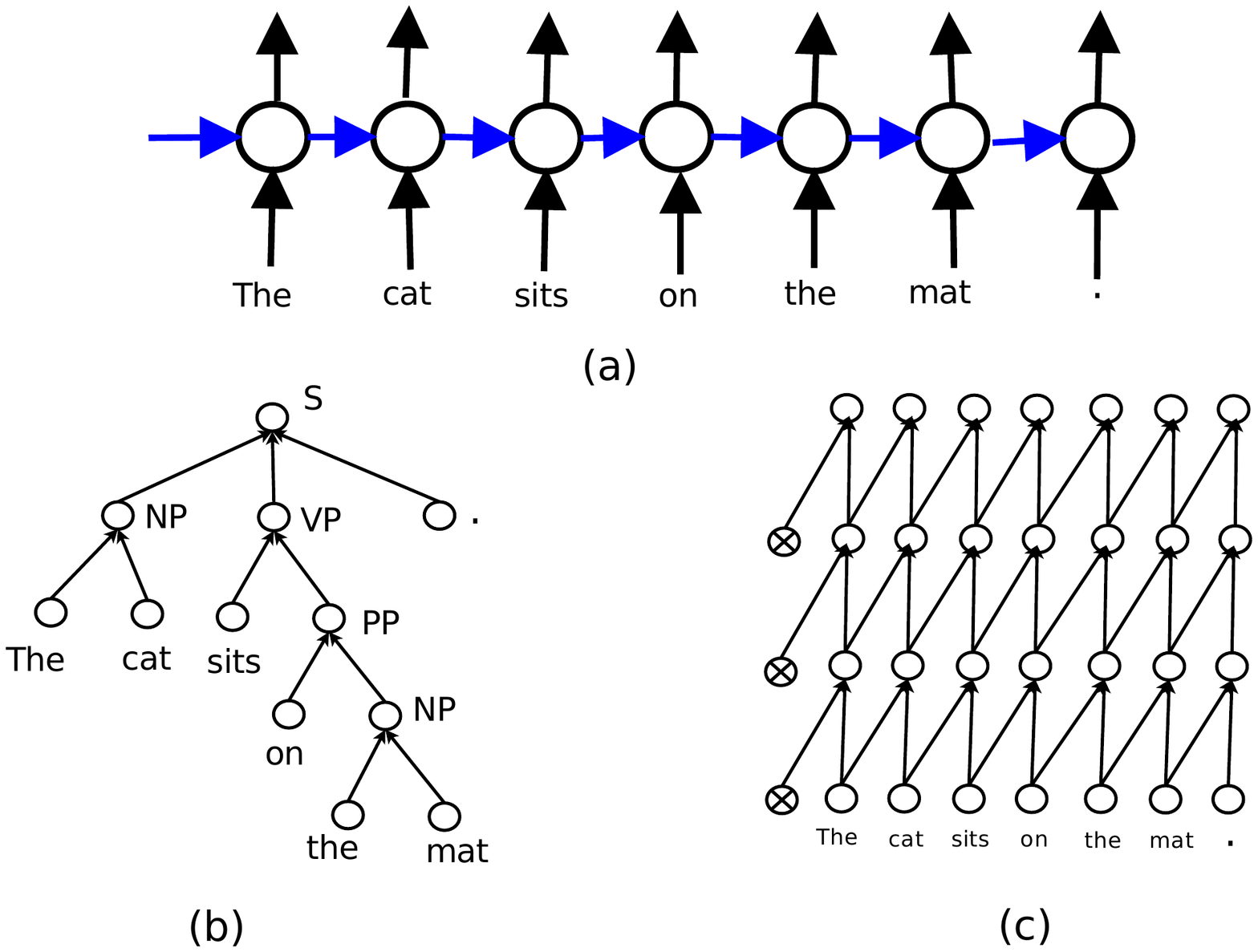}
\vspace{-0.5in}
\caption{\abc{Language models:} (a) RNN-based model, which takes the current word and the previous hidden states as input \cite{nic}; (b) the tree structure of a sentence, where \{S, NP, VP, PP\} denote sentence, noun phrase, verb phrase, and preposition phrase 
\cite{renn, cvg}; (3) CNN-based model, which is much denser than a tree structure, but is able to learn a tree structure \cite{mcnn}. $\otimes$ denotes zero padding.}\label{fig1}
\end{figure}

Human beings are able to describe what they see, and image captioning is a task that can make a computer have this ability. To achieve this goal requires at least three models: (1) vision model---to extract visual features from images, (2) language model---to generate captions, (3) the connection between vision and language models. Image captioning combines two fields---computer vision and natural language processing (NLP)
\abc{to address the challenge of understanding both images and their descriptions.}

Recently, increasing research has been devoted to image captioning, and a variety of methods have been proposed \cite{babytalk, milcap, mrnn, nic, att, sematt, whatvalue, scacnn, sental, scn, lstma}. Almost all of the current proposed methods are under the framework of CNN$+$RNN, in which a CNN is used for the vision model, and an RNN is employed to generate sentences. Moreover, there are different ways to connect the CNN and RNN. A naive way is directly feeding the output of the CNN into the RNN \cite{mrnn, nic}. However, this naive approach treats objects in an image the same and ignores the salient objects when generating one word. To imitate the visual attention mechanism of humans, attention modules have been introduced into the CNN$+$RNN framework \cite{att, sematt, sental}. 

Although the CNN$+$RNN framework is popular and provides satisfying results, there are disadvantages of sequential RNNs: (1) RNNs have to be calculated step-by-step, which is not amenable to parallel computing \abc{during training}; (2) there is a long path between the start and end of the sentence using RNNs (\abc{see \reffig{fig1}a}), \wqz{which easily forgets the long-range information}.
% \TODO{insert why this is bad}. 
Tree structures can make a shorter path between the start and end words (\abc{see \reffig{fig1}b}), but trees require special processing \abc{that is not easily parallelizable}. Also trees are defined by NLP (e.g., noun phrase, adjectives, verb phrase, etc), which is a hand-crafted structure that may not be optimal for the captioning task. 

An alternative language model to RNNs and trees are CNNs applied to the sentence to merge words layer by layer \cite{hierlstm, mcnn}, thus learning a tree structure of the sentence (\abc{see \reffig{fig1}c}). 
\abc{CNNs} can be implemented in parallel, and have a larger receptive field size (\abc{can see more words}) using less layers. E.g., given a sentence composed of 10 words, a RNN needs to iterate 10 times to get a representation of the sentence (equivalent to 10 layers), while a CNN with kernel size 3 only needs 5 layers. Using wider kernels, CNNs are able to tackle longer sentences with less layers.  
CNN have been widely applied to the field of NLP, leading to improved performances in many tasks, such as language modeling, machine translation and text classification \cite{glu, conv2conv, bytenet, emnlp2014cnn, eacl2017cnn}. 

Inspired by the applications of CNNs in the field of NLP, we develop a framework that only employ CNNs for image captioning. The main contributions of this paper are:
\begin{compactitem}
\item[1.] We propose a CNN$+$CNN framework for image captioning, which is faster than LSTM-based models and outperforms them on some metrics.
\item[2.] We propose a hierarchical attention module to connect the vision CNN  with the language CNN, which significantly improves the performance.
\item[3.] We investigate the influence of the hyper-parameters, including  
the number of layers and the kernel width of the language CNN.
The receptive field of the language CNN can be increased by stacking more layers or increasing the kernel width, and our experiments show that increasing the kernel width is better.
\end{compactitem}

\section{Related work}\label{sec2}
\paragraph{CNN$+$RNN models}

Many methods have been proposed to automatically generate captions of images, and the most popular framework is CNN$+$RNN. Most work %f work 
endeavors to modify the language model and the connection between the vision and language models to improve the performance.

The m-RNN model \cite{mrnn} uses a vanilla 
RNN combined with different CNNs (AlexNet \wqz{or} VGG-net)
% \NOTE{and or or? they use both at the same time?})
 -- the RNN hidden states and the CNN output are fed into a multimodal block to fuse the image and language features at each time step, and a softmax layer predicts the next word. However, the vanilla RNNs suffers from the vanishing gradient problem. Therefore, the neural image captioning (NIC) model employs a LSTM as a decoder \cite{nic}. At the beginning, the NIC model takes the image feature vector as input, and then the visual information is passed through the recurrent path.

In both m-RNN and NIC, an image is represented by a single vector, which ignores different areas and objects in the image. A spatial attention mechanism is introduced into image captioning model in \cite{att}, which allows the model to pay attention to different areas at each time step.

Most recently,  \cite{sematt, whatvalue, lstma} have included semantics, \abc{which are image annotations produced by an image classifier}, into image captioning.
A semantic attention model is proposed in \cite{sematt}, which applies a fully convolutional network (FCN) to detect semantics first and then computes a weight for each semantic at each time step. Although this semantic attention significantly improves the performance, to some extent, it has a problem of prediction error accumulation along generated sequences \cite{lstma}. Alternatively, \cite{scn} uses semantics differently, by generating the parameters of an RNN or LSTM from the output of the semantic detector. This model can be trained in a end-to-end manner.

LSTMs have also been used hierarchically to model a shallow tree structure. \cite{philstm} proposes a phrase LSTM model, which has two levels of LSTMs, one to model the sentence composed of phrases, and another to generate words in a phrase. Similarly, \cite{skeleton} applies a skeleton LSTM to generate the skeleton words of the sentence,  and an attribute LSTM to generate adjective words that modify the skeleton words, which is a coarse-to-fine model. Hierarchical LSTMs have also been used to generate a paragraph description consisting of multiple sentences to describe an image \cite{paralstm}.

%Currently, the most similar model is proposed by \cite{cnnl}, however, their model is also under the CNN$+$RNN framework. Instead of feeding one word into a RNN, \cite{cnnl} adopts a language CNN to extract features from the context, which is able to boost the long-range memory of RNNs.  

\wqz{To our knowledge, our work is the first that applies CNNs to completely replace RNNs for image captioning. 
%Although \cite{cnnl} uses language CNNs \abcn{to extract word contexts}, it is still under the RNN-based model \abcn{as it uses an LSTM to generate the captions.}
%, in which CNNs are adopted to extract features from the context. 
Inspired by \cite{mcnn}, we propose a hierarchical attention module, which aims to learn the co-relationship between concepts at each level and image areas. Furthermore, our attention module uses the dot-product, which has less parameters and can be calculated faster than the MLP-based attention in \cite{att, sental}.}

%\TODO{contrast with the CNN+RNN methods. is ours the first work to use CNN language model for captioning?  is our attention or hierarchical attention different than other attention models? do we use semantics?}

\paragraph{CNNs in NLP}
Deep learning methods have dominated the field of NLP, and CNNs are an important tool to solve NLP problems. In \cite{nlpjmlr}, CNNs are applied to several NLP tasks, such as chunking, part-of-speech tagging, named entity recognition, and semantic role labeling. This CNN-based model provides accurate results with fast speed, and is capable of learning representations instead of 
%Moreover, instead of 
employing hand-crafted features.

%Reference 
\cite{emnlp2014cnn} develops a CNN model with a max-over-time pooling layer for sentence classification. \cite{eacl2017cnn}  experiments with very deep CNNs for text classification, and the results suggest that depth and max-pooling improve the performance. 

In terms of language modeling, \cite{glu} introduces a new \wqz{activation} %\NOTE{activation?} 
function called gated linear units (GLU), \wqz{which applies an output gate to each unit, and can be trained faster than ReLUs}. % \NOTE{not sure what that means. Similar in what way? out of place sentence. what does it have to do with GLUs?}.
 In \cite{conv2conv}, GLUs are adopted for machine translation, where a convolutional  sequence-to-sequence model outperforms %is set up which is superior to 
the popular LSTM-based sequence-to-sequence model \cite{seq2seq, iclr2015mt}. 
%And i
In \cite{bytenet}, dilated convolution is employed to increase the receptive field of the CNN.
%CNNs are adopted as well and to increase the receptive field, dilated convolution is employed.

\wqz{In this paper, our proposed model adopts language CNNs without pooling layers, which is different from \cite{emnlp2014cnn, eacl2017cnn}. Furthermore,  we use {\em causal} convolution (similar to \cite{wavenet}), \abcn{so that our model can generate captions word-by-word.}}

%\TODO{any contrast with the CNNs for NLP?  Do we use a different structure, like non-centered (causal) conv windows?}

\section{Model}\label{sec3}

\begin{figure}[t]
\centering
\includegraphics[width=0.5\textwidth]{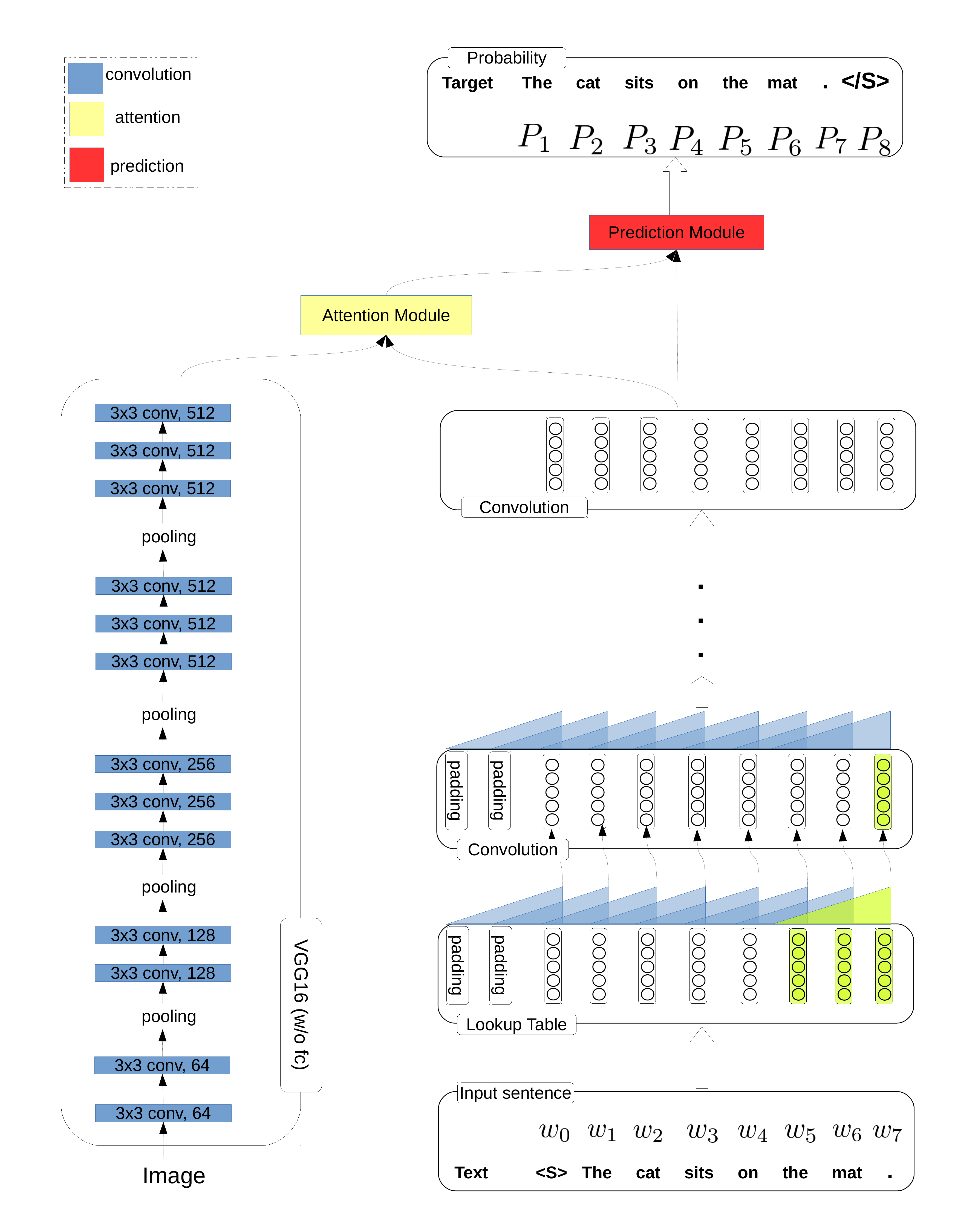}
\vspace{-0.3in}
\caption{The CNN$+$CNN model %framework that applies CNNs 
for image captioning. The vision CNN extracts features from the image, and the language CNN is applied to model the sentence.  The attention module and prediction module fuse the information from image and word contexts.
\abc{The convolutions in the language CNN use {\em causal} filters (depending only on the present and past words), to enable feed-forward sentence generation}. For each convolutional layer in the language module, we use $k-1$ zero vectors to pad the sentence matrix along the length axis, where $k$ denotes the kernel size (here, $k=3$).  
%\TODO{flip the diagram vertically, so that the "top-level" layer is at the top.}
%. In this figure we show the case $k=3$.
%\TODO{use "right" triangles for the language CNN convolutions.}
}
\label{fig2}
\end{figure}

\wqzp{CNNs show relatively strong abilities to tackle very long sequences} \cite{wavenet}.
% \NOTE{sentence doesn't parse}. 
Inspired by CNNs used for NLP, we propose a CNN$+$CNN frame work for image captioning. There are four modules in our framework: (1) vision module, which is adopted to ``watch'' images; (2) language module, which is to model sentences; (3) attention module, which connects the vision module with the language module; (4) prediction module, which takes the visual features from the attention module and concepts from the language module as input and predicts the next word. \reffig{fig2} illustrates the proposed framework for image captioning.

%\NOTE{I use $i$ for indexing the image position, and $j$ to index the sentence positions.}

\paragraph{The vision module}
The vision module is a CNN without fully connected layer, whose output is a $d\times d \times D_c$ feature map.
%, which has $d^2$ positions and each position corresponds to one area of the input image. 
%
In each position of the feature map, the $D_c$ dimensional feature vector represents part of the image.
Define $\textbf{v}=[v_{1},\cdots,v_{N}]$ to be the list of image feature vectors, where $v_{i} \in \mathbb{R}^{D_c}$, $i$ is the position index in the feature map, and $N=d^2$ is the number of positions.  In this paper, we use VGG-16 \cite{vgg} as the CNN for the vision module.

\paragraph{The language module}
%The language module are very different from the RNN-based framework, where a CNN without pooling and fully connected layers is adopted. 
The language model is based on a CNN without pooling, which is very different from the typical RNN-based framework (e.g., \cite{mrnn, nic, lrcn, att}). % \TODO{insert citations}).
RNNs adopt a recurrent path to memorize the context, while CNNs use kernels and stack multiple layers to model the context.

%A sentence constituted by $L$ words can be represented as 
Let $S=[w_1, w_2, \cdots, w_{L}]$ be a sentence with $L$ words. We first use a look-up table to project each word into an embedding space with $D_e$ dimensions, and calculate the embeddings $E=[e_1, e_2, \cdots, e_{L}]$, where $e_j\in \mathbb{R}^{D_e}$. In this stage, a sentence is represented by a $L \times D_e$ matrix. A stack of convolutional layers with gated linear units (GLUs) follows the embedding layer, which is calculated using the following equations:
\begin{align}
h_a^l &= W_a^l*h^{l-1} + b_a^l, \\
h_b^l &= W_b^l*h^{l-1} + b_b^l,\\
h^l &= h_a^l \odot \sigma(h_b^l),
\end{align}
where $W_a^l, W_b^l \in \mathbb{R}^{k\times D_e}$ denote the kernels of the $l$th layer,  $b_a^l$ and $b_b^l$ are biases, $*$ denotes the convolution operator, $\odot$ denotes the element-wise multiplication, and $\sigma(x)=\frac{1}{1+e^{-x}}$. $h_b^l$ plays the role of a gate and $h_a^l$ is a linear transformation. In our framework $h^0=E$.
\abc{Note that the convolution filters are {\em causal} filters, only depending on the current and past inputs.  During inference, this structure allows a sentence to be generated using a feed-forward process where the predicted word at the output layer is used as the next word input.}

For modeling sentences, the length of the output is required to be the same as the input sentence. 
Since there are no pooling layers and fully connected layers, we need to add zero-padding at the beginning of the input and hidden layers.
%the CNN is able to tackle the variable lengths of sentences. 
 If the convolutional kernel width is $k$, which indicates that it considers $k$ concepts\footnote{The bottom-level represents words, whereas hidden layers represent different concepts.} in each step, the word/concept matrices needs to be zero padded with $k-1$ zero vectors before convolution.
 %the output of a convolutional layer is $L-k+1$ concepts. To guarantee our model outputs $L$ concepts after convolution, we use $k-1$ zero vectors to pad the input, thus the input becomes 
%  a $(L+k-1)\times D_e$ matrix and each convolutional layer 
The output of the CNN is a set of concepts $\textbf{c}=[c_1, c_2, \cdots, c_{L}]$, where $c_j\in \mathbb{R}^{D_e}$.

\begin{figure}[t]
\centering
\includegraphics[width=0.5\textwidth]{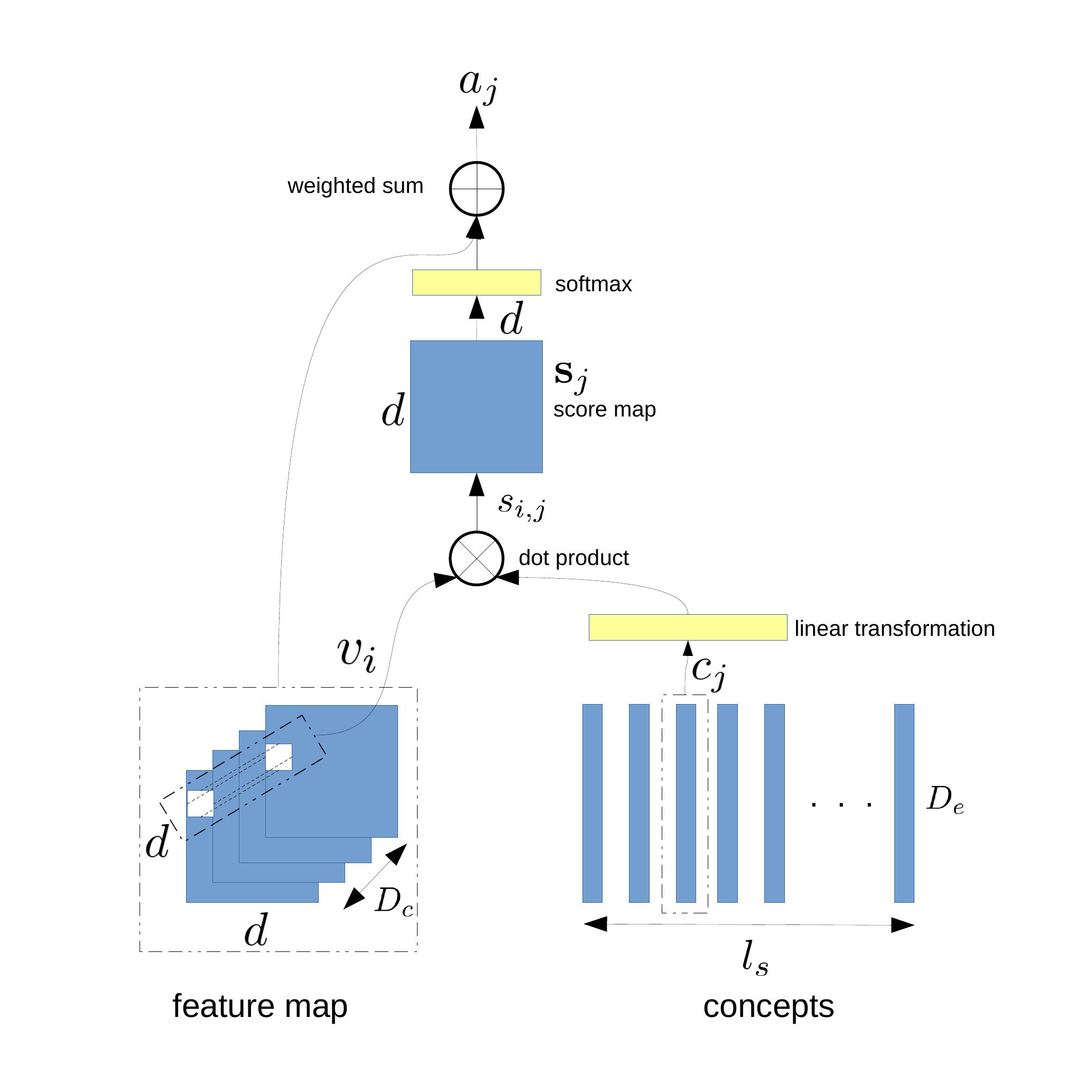}
\vspace{-0.4in}
\caption{The attention module. \wqz{We apply dot-product and soft-max to calculate the weight for each position \abcn{and concept}, which only has one parameter matrix (linear transformation). 
%While in \cite{att, sental}, the attention module contains 3 parameter matrices, which is more complicated and slower.
}
%\TODO{add more description.}
%\TODO{change "i,j" back to "i"...sorry changed my mind.  Follow the notation of the text.}
}\label{fig3}
\end{figure}

\paragraph{The attention module}
Intuitively, to predict different words, different objects in the images should be attended \abc{and input into the prediction module.} 
%Therefore we design this attention module to make our model pay attention to different areas.
The attention module takes the visual features $\textbf{v}$ and the concepts $\textbf{c}$ at the top level as input, which is shown in \reffig{fig3}. For each concept $c_j$ and visual feature vector $v_{i}$, we calculate a score $s_{i,j}$ as follows:
\begin{align}
s_{i,j} = c_j^TUv_{i},
\label{eqn:att_score}
\end{align}
where $U\in \mathbb{R}^{D_e\times D_c}$ is a \abc{parameter matrix}. Therefore each concept $c_j$ corresponds to a score vector $\textbf{s}_j = [s_{1,j}, s_{2,j}, \cdots, s_{N,j}]$, indicating matches with the image features. Then $\textbf{s}_j$ is fed into a softmax layer, which assigns a weight $w_{i,j}$ for the visual feature vector $v_{i}$,
\begin{align}
w_{i,j} = \frac{e^{s_{i,j}}}{\sum_{i=1}^{N}e^{s_{i,j}}}.
\label{eqn:att_weight}
\end{align} 
%The value of $w_{i,j}$ indicates how much attention the model should pay to the image area $i$ for concept $j$. 
We use the weighted sum operation to calculate the final attention vector as follows,
\begin{align}
a_j=\sum_{i=1}^{N} w_{i,j}v_{i}.
\end{align}
The output of our attention module is the attention features $\textbf{a} = [a_1,a_2,\cdots,a_{L}]$, where $a_i\in \mathbb{R}^{D_c}$, corresponding to each word in the sentence.

\paragraph{The prediction module}
\begin{figure}[t]
\centering
\includegraphics[width=0.5\textwidth]{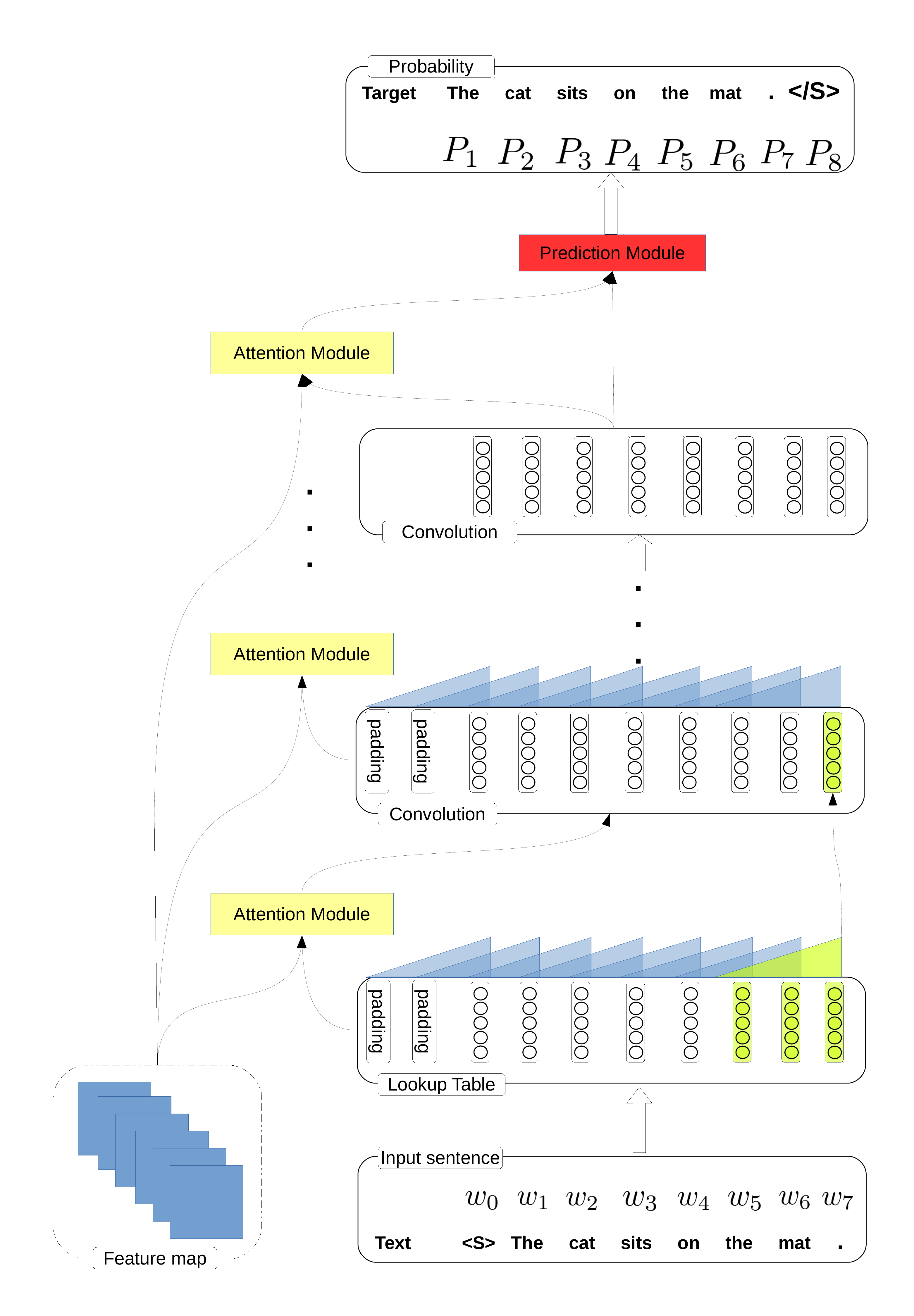}
\vspace{-0.3in}
\caption{The hierarchical attention model. \wqz{The structure of hierarchical attention moves up the layers, not to the right, so as to prevent sideways connections in the same layer (as in an RNN).}
%\TODO{add more description.}
%\TODO{change notation to follow text.}
%\NOTE{this figure could be cut to save space.}
}\label{fig4}
\end{figure}

%\NOTE{avoid using "time" since it is not an RNN...might confuse the reader.}
As shown in \reffig{fig2}, our proposed prediction module takes the attention features $\textbf{a}$ and concepts $\textbf{c}$ as input. Our prediction module is a one-hidden layer neural network.
At the $j$th sentence position, $a_j$ and $c_j$ are fed into the network, the output is the prediction probability of next word $P_{j+1}$,  %which is calculated as follows: 
\begin{align}
h^p_j &= f\left(W_a^pa_j + W_c^pc_j+ b^p\right), \\
P_{j+1} &= \mathop{\mathrm{softmax}}(U^ph^p_j),
\end{align}
where %$h^p_j$ denotes the hidden state of the prediction module at sentence position $j$, 
$W^p_a\in \mathbb{R}^{D^p_h\times D_c}$ and $W^p_c\in \mathbb{R}^{D^p_h\times D_e}$ are the parameters, and $D^p_h$ denotes the number of hidden units in the prediction module. $f(x)=\max(x, 0.1x)$ is a leaky ReLU, and $U^p\in \mathbb{R}^{V\times D^p_h}$, where $V$ represents the vocabulary size.
%
%The process is shown in \reffig{fig4}. 
During training the input image-sentence pair is given, therefore the prediction module can be implemented as a $1\times 1$ convolutional kernel, % to calculate the probabilities, 
which is  faster than RNN-based models.

\subsection{Hierarchical attention}
%We have demonstrated our basic CNN$+$CNN model for image captioning, where we calculate the attention features using the concepts at the top level of the language module, however, different levels of CNNs have different receptive fields, therefore the concepts are variable.
The basic CNN$+$CNN model in the previous sub-section extracts attention features from the top-level of the language model.
However, different levels of the language CNN represent different concept levels that could benefit from visual inputs.
Hence, we also consider a hierarchical attention (hier-att) module (\wqz{see \reffig{fig4}}), where attention vectors are computed at each level of the language model and fed into the next level, %decoder and feed them into the next level:
\begin{align}
a^{l-1}_j &= \sum_{i=1}^N w^{l-1}_{i,j}v_{i}, \\
h_a^l &= W_a^l*h^{l-1} + W_a^{att}*\textbf{a}^{l-1} + b_a^l, \\
h_b^l &= W_b^l*h^{l-1} + W_b^{att}*\textbf{a}^{l-1} + b_b^l, \\
h^l &= h_a^l \odot \sigma(h_b^l), 
\end{align}
where $w^{l-1}_{i,j}$ is computed using equations (\ref{eqn:att_score}-\ref{eqn:att_weight}) \abc{on the concepts $\mathbf{c}^{l-1} $at level $l-1$}, $W_a^{att},W_b^{att}\in \mathbb{R}^{D_e\times D_c}$ and $\textbf{a}^{l-1}=[a^{l-1}_j]_{j=1}^{L}$.

\wqz{In contrast to RNN-based models that calculate attention maps in a left-right (word-by-word) manner, 
our attention maps are calculated in a bottom-up (layer-by-layer) manner, so as to prevent sideways connections in the same layer. This allows our model to be trained in parallel over all words in the sentence, rather than word-by-word.}  
\abcn{Note that the model still sees attended features of previous words (from the lower layer) through the causal convolution layer.}

%\TODO{need a figure to illustrate hierarchical attention}
%\TODO{we have a specific structure of hierarchical attention that moves up the levels, not to the right, so as to prevent sideways connections in the same layer (as in an RNN).  You should mention it.}

\subsection{Training and inference}
During training, as image-caption pairs are given, the convolution structures are applied %we can apply convolution 
in the normal way, and \abc{the loss function for each sentence is the  cross-entropy,}
\begin{align}
{\cal L} =  -\sum_{j=0}^{L-1} y_{j+1}\log(P_{j+1}) + \frac{\lambda}{2}\sum_W||W||^2, 
\end{align}
where $y_{j+1}$ denotes the ground-truth label and $P_{j+1}$ is the prediction probability. The first term is the same with the loss function in \cite{nic}, and the second term is the regularization term, where $W$ denotes the weights in our framework.

\abc{During inference, the caption is generated given the image using a feed-forward process. The  caption is initialized as zero padding and a start-token $<$S$>$, and is fed as the input sentence to the model to predict the probability of the next word.  The predicted word is appended to the caption, and the process is repeated until the ending token $<$/S$>$ is predicted, or the maximum length is reached. 
The predicted words are selected using a naive greedy method (most probable word), which is equivalent to a beam-search algorithm with beam width of 1.}

%we apply a naive way which is equivalent to the beam search algorithm with beam width $bw=1$. Our model starts with the starting token $<S>$ and stop when it predicts the ending token $</S>$ or meets the maximum length.
 
%\NOTE{did you try actual beam search? does it improve the results?}
%\TODO{add figure if space}
 
\section{Experiments}

We present experiments using our proposed CNN$+$CNN model on three datasets.

\subsection{Dataset and experimental setup}

\paragraph{MSCOCO} MSCOCO is the most popular dataset for image captioning, comprising  82,783 training and 40,504 validation images. Each image has 5 human annotated captions. As in \cite{deepvs, att}, we split the images into 3 datasets, consisting of 5,000 validation and 5,000 testing images, and \wqz{113,287 training images}. %\TODO{how many for training?}. 
Our vocabulary contains 10,000 words.

\paragraph{Flickr30k} Flickr30k contains 31,783 images, each of which has 5 captions. Following \cite{deepvs}, %\TODO{INSERT ref}, 
we use 1,000 images for validation and 1000 images for testing, and the remainder for training. The vocabulary size is 10,000.

\paragraph{Paragraph annotation dataset (PAD)} A dataset with 19,551 images from MSCOCO and visual genome (VG). Each image is described by a paragraph comprising several sentences. The average length of the paragraph is 67.5 words, while that of MSCOCO dataset is 11.3. Generating such long captions is much more challenging. %In this paper
We use the same split as \cite{paralstm}, and our vocabulary size is 5,000.

\paragraph{Implementation details}For all datasets, we set $D_e=300$. In our framework, the vision module is the VGG-16 network without fully connected layers, which is pre-trained on ImageNet.\footnote{\url{https://github.com/tensorflow/models/tree/master/research/slim}} We use $\mathrm{conv}5\_3$ feature map to compute attention features ($D_c = 512$ and $d=14$). The size of the hidden layer in the prediction module is $1024$, and $\lambda$=1e-5. We apply Adam optimizer \cite{adam} with mini-batch size 10 to train our model. For VGG-16, we set the learning rate to 1e-5, % $0.00001$,
 and for other parameters, the learning rate starts from 1e-3 %0.001 
 and decays every 50k steps.\footnote{For PAD, the learning rate decays every 10k steps.} The stopping criteria is based on the validation loss. % the validation dataset. 
The weights are initialized by the truncated normal initializer with $stddev=0.01$. The model was implemented in TensorFlow.

During training time, the images are resized to $299\times 299$ and then we randomly crop $224\times 224$ patches as inputs. During testing time, the images are resize into $224\times 224$ without cropping.\footnote{We employ the code provided by the following link: \url{https://github.com/tensorflow/models/tree/master/research/im2txt}}  The maximum caption length is 70 for MSCOCO and Flickr30k, and 150 for PAD. 
%paragraph annotation dataset.

\paragraph{Compared models} We compare our proposed model with and without hierarchical attention.
% and the model with hierarchical attention. Moreover, 
We also compare other popular models: DeepVS \cite{deepvs}, m-RNN \cite{mrnn}, Google NIC \cite{nic}, LRCN \cite{lrcn}, hard-ATT and soft-ATT \cite{att} %, $\text{CNN}_L$ \cite{cnnl} and $\text{CNN}_L+\text{LSTM}$ \cite{cnnl} 
on MSCOCO and Flickr30k datasets, and Sentence-Concat, DenseCap-Concat, Image-Flat, Regions-Flat-Scratch, Regions-Flat-Pretrained, and Regions-Hierarchical \cite{paralstm} on PAD. %paragraph annotation dataset.

\paragraph{Metrics} We report the following metrics: \textbf{B}LEU-1,2,3,4 \cite{bleu}, \textbf{M}eteor \cite{M}, \textbf{R}ouge-L \cite{R} and \textbf{C}IDEr \cite{C}, which are computed using the MSCOCO toolkit \cite{toolkit}. For all metrics, higher values indicate better performance.
 % value is higher, the performance is better.

%\subsection{Results}

%%%%%%%%%%%%%%%%%%%%%%%%%%%%%%%%%%%%%%%%%%%%%%%%%%%%%COCO%%%%%%%%%%%%%%%%%%%%%%%%%%%%%%%%%%
\begin{table*}[t]
\begin{center}
\small
\begin{tabular}{c|c|c|c|c|c|c|c|c}

\hline
\textbf{Dataset} &\textbf{Model} &\textbf{B-1} &\textbf{B-2} &\textbf{B-3} &\textbf{B-4} &\textbf{M} &\textbf{R} &\textbf{C} \\

\hline
\multirow{8}{*}{MSCOCO}
&DeepVS \cite{deepvs} &0.625 &0.450 &0.321 &0.230 &0.195 &- &0.660\\

\cline{2-9}
&m-RNN \cite{mrnn} &0.670	&0.490	&0.350	&0.250 &- &- &- \\

\cline{2-9}
&NIC \cite{nic} &0.666	&0.461	&0.329	&0.246	&-	&-	&- \\

\cline{2-9}
&LRCN \cite{lrcn} &\textcolor{blue}{0.697} &\textbf{0.519} &\textbf{0.380} &\textbf{0.278} &0.229 &\textcolor{red}{0.508} &\textcolor{blue}{0.837}\\

\cline{2-9}
&Hard-ATT \cite{att} &\textbf{0.718} &0.504 &0.357 &0.250 &\textcolor{blue}{0.230} &- &-\\

\cline{2-9}
&Soft-ATT \cite{att} &\textcolor{red}{0.707} &0.492 &0.344 &0.243 &\textbf{0.239} &- &-\\

%\cline{2-9}
%&$\text{CNN}_L$ \cite{cnnl} &- &- &- &0.184 &- &- &0.568 \\

%\cline{2-9}
%&$\text{CNN}_L+\text{LSTM}$ \cite{cnnl} &\textbf{0.721} &\textbf{0.546} &\textbf{0.409} &\textbf{0.304} &\textbf{0.251} &- &\textbf{0.991} \\

\cline{2-9}
&Ours (w/o hier-att) &0.688 &\textcolor{red}{0.513} &\textcolor{red}{0.370} &\textcolor{blue}{0.265} &\textcolor{red}{0.234} &\textcolor{blue}{0.507} &\textcolor{red}{0.839} \\

\cline{2-9}
&Ours (w/ hier-att) &0.685 &\textcolor{blue}{0.511} &\textcolor{blue}{0.369} &\textcolor{red}{0.267} &\textcolor{red}{0.234} &\textbf{0.510} &\textbf{0.844} \\

\hline
\hline

\multirow{8}{*}{Flickr30k}
&DeepVS \cite{deepvs} &0.573 &0.369 &0.240 &0.157 &0.153 &- &\textcolor{blue}{0.247}\\

\cline{2-9}
&m-RNN \cite{mrnn} &0.60 &0.41 &0.28 &0.19 &- &- &- \\

\cline{2-9}
&NIC \cite{nic} &\textcolor{blue}{0.663} &0.423 &0.277 &0.183	&-	&-	&- \\

\cline{2-9}
&LRCN \cite{lrcn} &0.587 &0.39 &0.25 &0.165 &- &- &- \\

\cline{2-9}
&Hard-ATT \cite{att} &\textbf{0.669} &\textbf{0.439} &\textbf{0.296} &\textbf{0.199} &\textcolor{red}{0.185} &- &-\\

\cline{2-9}
&Soft-ATT \cite{att} &\textcolor{red}{0.667} &\textcolor{red}{0.434} &\textcolor{blue}{0.288} &\textcolor{red}{0.191} &\textcolor{red}{0.185} &- &-\\

%\cline{2-9}
%&$\text{CNN}_L$ \cite{cnnl} &- &- &- &- &- &- &- \\

%\cline{2-9}
%&$\text{CNN}_L+\text{LSTM}$ \cite{cnnl} &\textcolor{blue}{0.645} &\textbf{0.458} &\textbf{0.322} &\textbf{0.224} &\textcolor{red}{0.190} &- \\

\cline{2-9}
&Ours, w/o hier-att &0.577 &0.401 &0.276 &\textcolor{blue}{0.190} &\textcolor{blue}{0.184} &\textcolor{red}{0.425} &\textcolor{red}{0.352}\\

\cline{2-9}
&Ours, w/ hier-att &0.607 &\textcolor{blue}{0.425} &\textcolor{red}{0.292} &\textbf{0.199} &\textbf{0.191} &\textbf{0.442} &\textbf{0.395} \\

\hline

\end{tabular}
\end{center}
\caption{Performance on MSCOCO and Flickr30k dataset. B, M, R, C stand for BLEU, Meteor, Rouge-L and CIDEr, respectively. The bold, red, and blue numbers are the highest value, 2nd highest and 3rd highest values, respectively.}\label{tab1}
\end{table*}

\subsection{Results on MSCOCO and Flickr30k}

\begin{figure}[t]
\centering
\includegraphics[width=0.5\textwidth]{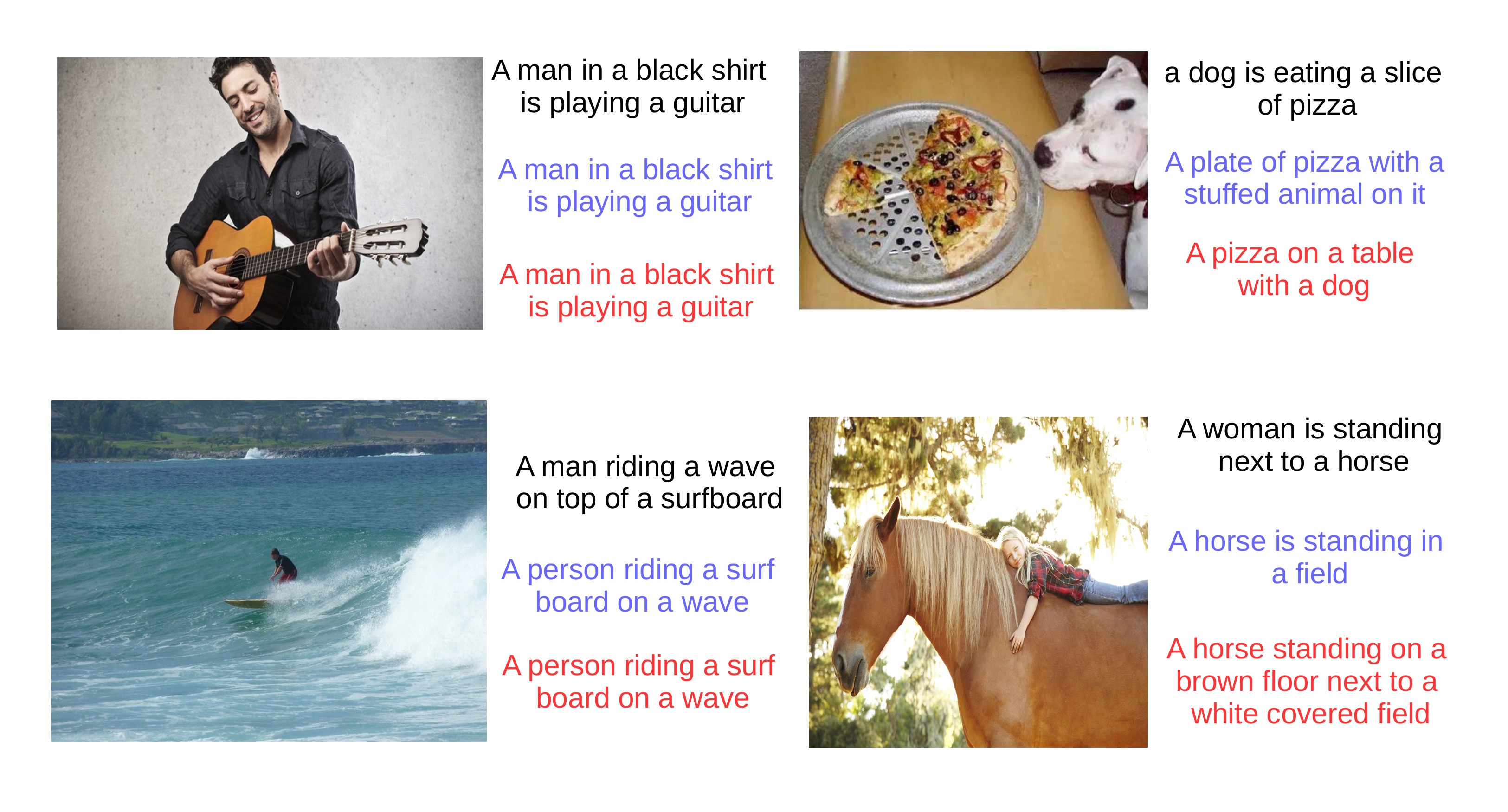}

\caption{Examples of generated descriptions. 
Black, blue and red text correspond to captions from NIC, our model, and our model with hier-att.
}\label{example}
\end{figure}

Table \ref{tab1} shows the results on MSCOCO and Flickr30k. Note that all models only use the image information without semantics or attributes boosting. In terms of the B-1 score, which only considers bigrams, our model performs slightly worse than Hard- and Soft-ATT models, but better than NIC and m-RNN on MSCOCO. In terms of B-2,3,4 and M metrics, the performance of our model ranks 2nd, and the values are marginally lower than the LRCN model, which applies multilayer LSTMs to generate captions. %However, when compared with $\text{CNN}_L+\text{LSTM}$ model, there is a gap between our model and $\text{CNN}_L+LSTM$, but our model performs much better than $\text{CNN}_L$, 
This suggests  that  adopting CNNs as decoder is competitive or can be better than using LSTMs. When we apply our proposed hierarchical attention model, the B-4, M, R and CIDEr scores exhibit improvement. \wqz{Using hier-att with our model improves the CIDEr score from 0.839 (no hier-att) to 0.844 and the Meteor score from 0.507 to 0.510.} 

On Flickr30k, our proposed model also provides the comparable results, and can slightly improve the M score compared with LSTM-based models. Moreover, our proposed hierarchical attention significantly improves all scores compared with the model without hierarchical attention. \wqz{The CIDEr score is improved from 0.352 to 0.395 (12\% increase), B-1 score is improved by 3\% from 0.507 to 0.607 and other scores are improved by around 1\%.}

An advantage of our proposed CNN$+$CNN model is the training speed. We compare our 6-layer model without hierarchical attention with NIC. When we set the batch size to 32, and train our model and NIC  on the same machine with a GTX1080 GPU, it takes 0.24 seconds per batch for our model, and 0.64 seconds for NIC. % per \wqz{batch} for our model and NIC model, respectively.  
Our model is about $3\times$ faster than NIC for training.\footnote{NIC is the simplest among the LSTM-based models.} \wqz{After training 50k steps (each step uses one batch), the loss decreases by around 70\%, and the model is able to output reasonable descriptions.}
%Moreover, when we apply hierarchical attention or increase the layers of the decoding CNN, the training time is also acceptable.
 In the inference stage, both our CNN$+$CNN model and LSTM-based model generate one word at a time, and thus, the inference speed is almost the same. 

\wqz{Note that the NIC model contains about 6.7M parameters, while our proposed model contains more than 16.0M parameters.}
\abcnn{Although our model is more complex, it can be trained faster because it can better take advantage of parallelization since there are no recurrent (left-right) connections between words within the same layer.} \wqzpp{Examples of generated descriptions  are shown in \reffig{example}.}
%We also did not see any overfitting 
%which is much more complicated, but overfitting does not occur 
%on MSCOCO 
%when we apply a 6-layer CNN to generate captions.} 

%\NOTE{any hard numbers for the hier-att model?}
%\NOTE{how does NIC compare with other models? is it the fastest among LSTM-based models?}
%\NOTE{how about comparison on the number of parameters (not including the visual module)?}

%\TODO{add some example image/captions from MSCOCO and Flick. Show captions from a few models.}

\subsection{Results on paragraph generation} 

%%%%%%%%%%%%%%%%%%%%%%%%%%%%%%%%%%% VG %%%%%%%%%%%%%%%%%%%%%%%%%%%%%%%
\begin{table*}[t]
\begin{center}
\small
\begin{tabular}{c|c|c|c|c|c|c|c}
\hline
\textbf{Model} &\textbf{B-1} &\textbf{B-2} &\textbf{B-3} &\textbf{B-4} &\textbf{M} &\textbf{R} &\textbf{C} \\

\hline
$\text{Sentence-Concat}^{*}$   &0.311 &0.151 &0.076 &0.040 &0.120 &- &0.068 \\

\hline
$\text{DenseCap-Concat}^{*}$  &0.332 &0.169 &0.085 &0.045 &0.127 &- &0.125 \\

\hline
$\text{Image-Flat}^{*}$ \cite{deepvs}  &0.340 &0.200 &0.122 &0.077 &0.128 &- &0.111 \\

\hline
$\text{Regions-Flat-Scratch}^{*}$ &0.373 &0.217 &0.131 &0.081 &0.135 &- &0.111 \\

\hline
$\text{Regions-Flat-Pretrained}^{*}$ &0.383 &0.229 &0.142 &\textbf{0.090} &0.142 &- &0.121 \\

\hline
$\text{Regions-Hierarchical}^{*}$ &\textbf{0.419} &\textbf{0.241} &\textbf{0.142} &0.087 &\textbf{0.160} &- &0.135 \\

\hline
Ours, 20-layer, $k=3$ &0.350 &0.202 &0.118 &0.068 &0.140 &0.263 &0.106 \\

\hline
Ours, 20-layer, $k=5$ &0.311 &0.171 &0.099 &0.057 &0.127 &0.263 &0.120 \\

\hline
Ours, 20-layer, $k=7$ &0.350 &0.194 &0.107 &0.059 &0.133 &0.259 &\textbf{0.152} \\

\hline
\end{tabular}
\end{center}
\vspace{-0.1in}
\caption{Results on the paragraph annotation dataset. The bold numbers represent the highest score. We use a 20-layer CNN with skip connections every 3 layers, and we do not use hierarchical attention for this task. $^*$The results are from \cite{paralstm}. % In our experiments, we use a skip connection every 3 layers.
%\TODO{round the numbers to 3 digits (e.g., 0.123). Also add citations for the reference methods.}
}\label{tab2}
\end{table*}

\begin{figure}[t]
\centering
\includegraphics[width=0.5\textwidth]{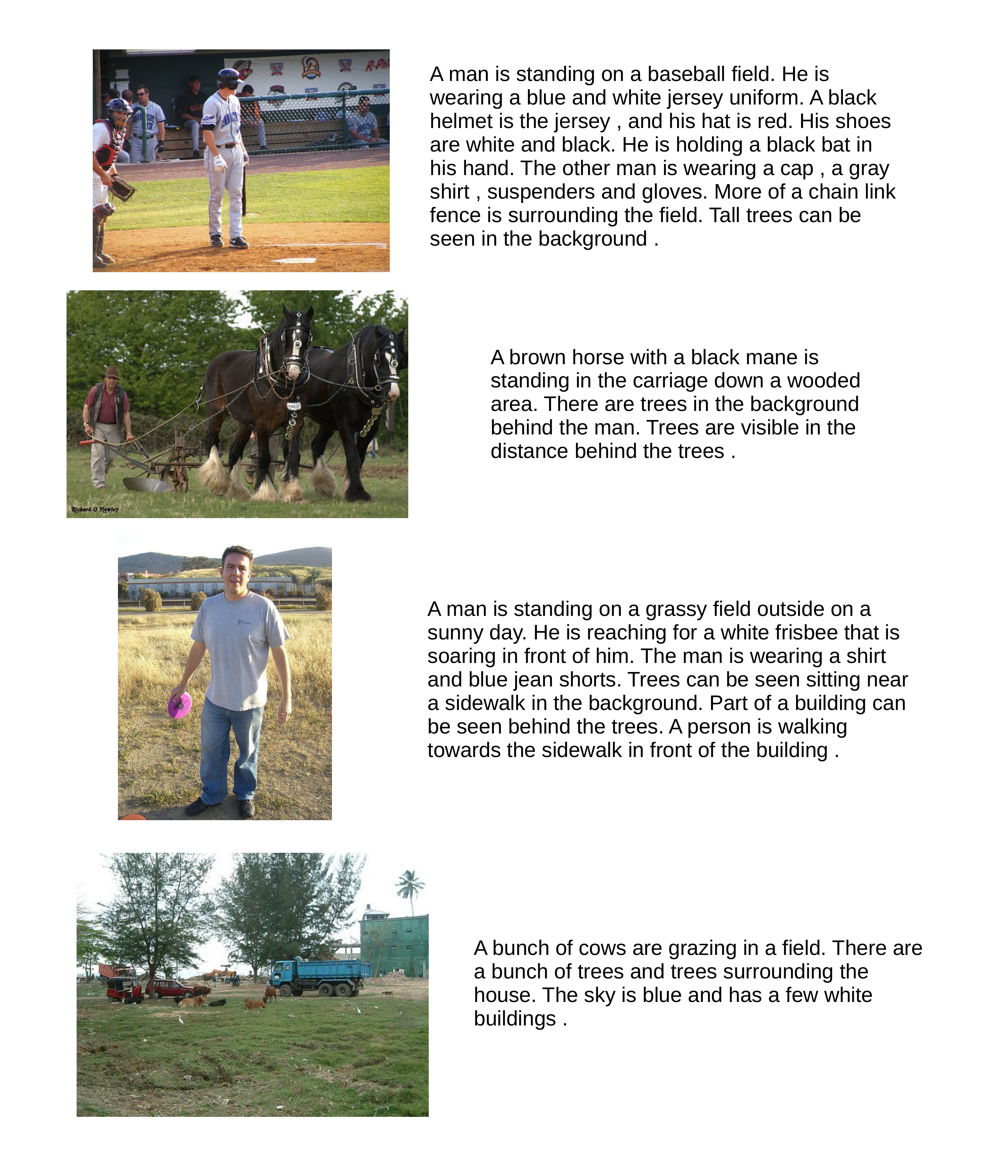}
\vspace{-0.4in}
\caption{Examples of generated paragraph description.}\label{fig5}
\end{figure}

We evaluate our model on the paragraph annotation dataset, in which an image has a much longer caption than MSCOCO and Flickr30k. Table \ref{tab2} shows the results. Our proposed model exhibits comparable performance with Image-Flat when considering B-1,2,3,4, which is lower than the Regions-Hierarchical.  Our model with kernel size 7 significantly improves the CIDEr score, which is specifically designed to evaluate image captioning models. Some examples generated by our 20-layer model with kernel size 7 are shown in \reffig{fig5}.

\subsection{Wider or Deeper?}

%%%%%%%%%%%%%%%%%%%%%%%%%%%%%%%%%%%%%%%%%%%%%%%%%%%%% ablative

\begin{table*}[t]
\begin{center}
\small
\begin{tabular}{c|c|c|c|c|c|c|c|c|c|c}
\hline
\textbf{Dataset} &\#\textbf{layers} &$k$ &\#\textbf{para} &\textbf{B-1} &\textbf{B-2} &\textbf{B-3} &\textbf{B-4} &\textbf{M} &\textbf{R} &\textbf{C} \\

\hline
\multirow{8}{*}{MSCOCO}
&\multirow{4}{*}{6, w/o hier-att}
&2 &16.0 M &0.679 &0.504 &0.363 &0.262 &0.231 &0.505 &0.827\\

\cline{3-11}
& &3 &16.9 M &0.683 &0.509 &0.367 &0.264 &0.234 &0.507 &0.829 \\

\cline{3-11}
& &5 &18.7 M &\textcolor{red}{0.687} &0.511 &0.368 &0.264 &0.233 &0.509 &0.838 \\

\cline{3-11}
& &7 &20.5 M &\textbf{0.688} &\textcolor{red}{0.513} &\textcolor{red}{0.370} &0.265 &0.234 &0.507 &0.839 \\

\cline{2-11}
&\multirow{4}{*}{6, w/ hier-att} 
&2 &16.0 M &0.682 &0.508 &0.368 &0.266 &0.233 &0.508 &0.836\\

\cline{3-11}
& &3 &16.9 M &0.684 &\textbf{0.514} &\textbf{0.372} &\textbf{0.269} &\textbf{0.235} &\textbf{0.511} &\textcolor{red}{0.842} \\

\cline{3-11}
& &5 &18.7 M &0.685 &0.511 &0.369 &\textcolor{red}{0.267} &0.234 &\textcolor{red}{0.510} &\textbf{0.844} \\

\cline{3-11}
& &7 &20.5 M &0.686 &0.510 &0.367 &0.264 &0.234 &\textcolor{red}{0.510} &0.834 \\
\cline{2-11}
&20 w/o hier-att &3 &24.5 M &0.683 &0.505 &0.365 &0.264 &0.234 &0.505 &0.838 \\
\hline
\hline

\multirow{8}{*}{Flickr30k}
&\multirow{4}{*}{6, w/o hier-att}
&2 &16.0 M &0.552 &0.378 &0.257 &0.175 &0.179 &0.421 &0.342\\

\cline{3-11}
& &3 &16.9 M &0.577 &0.396 &0.267 &0.181 &0.180 &0.426 &0.350 \\

\cline{3-11}
& &5 &18.7 M &0.577 &0.396 &0.268 &0.183 &0.183 &0.424 &0.340 \\

\cline{3-11}
& &7 &20.5 M &0.577 &0.401 &0.276 &0.190 &0.184 &0.425 &0.352 \\

\cline{2-11}
&\multirow{4}{*}{6, w/ hier-att} 
&2 &16.0 M &0.581 &0.403 &0.277 &0.190 &0.182 &0.433 &0.368\\

\cline{3-11}
& &3 &16.9 M &\textbf{0.607} &\textbf{0.425} &\textbf{0.292} &\textbf{0.199} &\textbf{0.191} &\textbf{0.442} &\textbf{0.395} \\

\cline{3-11}
& &5 &18.7 M &0.588 &0.402 &0.270 &0.185 &0.183 &0.428 &0.352 \\

\cline{3-11}
& &7 &20.5 M &\textcolor{red}{0.593} &\textcolor{red}{0.412} &0.282 &0.193 &\textcolor{red}{0.187} &\textcolor{red}{0.437} &\textcolor{red}{0.386} \\
\cline{2-11}
&20 w/o hier-att &3 &24.5 M &0.589 &0.411 &\textcolor{red}{0.284} &\textcolor{red}{0.198} &0.175 &0.425 &0.320 \\
\hline
\end{tabular}
\end{center}
\vspace{-0.08in}
\caption{The influence of the  kernel width and number of layers. \#layers denotes the number of layers, $k$ denotes the width of kernel, and \#para is the number of parameters of the model, which reflects the complexity of the model. 
%\NOTE{use consistent colors for best, 2nd best, 3rd best.}
%B, M, R, C stand for BLEU, Meteor, Rouge-L and CIDEr, respectively. The bold numbers represent the best ones and the blue numbers are the second best.
}\label{tab3}
\end{table*}

In this experiment, we compare the performance of our model with different kernel widths and depths, with results shown in Table \ref{tab3}. 
Designing and training deeper networks is popular, yet in our experiments on image captioning it is unnecessary to use deeper network \abc{on these datasets.} The results %shown in table \ref{tab3} 
suggest that for MSCOCO and Flickr30k datasets 6-layer networks with kernel size 3, and hierarchical attention is the optimal choice. 

For our proposed model without hierarchical attention, increasing the kernel width is a better choice both on MSCOCO and Flickr30k. When the kernel size increases from 2 to 7, the CIDEr score increases from 0.827 to 0.839 on MSCOCO, and from 0.342 to 0.352 on Flickr30k, which is better than the 20-layer network. 

Note that the average length of the captions in MSCOCO is 11.6, and the receptive field of a 6-layer CNN with kernel size 3 is 13, which is longer than many sentences in MSCOCO. Increasing the kernel width  or the network depth increases the complexity of the model, which makes it is more difficult to train and easier to overfit. Based on our experiments, we suggest using $k=3$ and setting the depth such that the receptive field is %to 
the average length of the sentences, which gives relatively good balance between the complexity and model performance.
\vspace{-0.1in}
\subsection{Attention module analysis}
\vspace{-0.05in}

\begin{figure}
\centering
\includegraphics[width=0.5\textwidth]{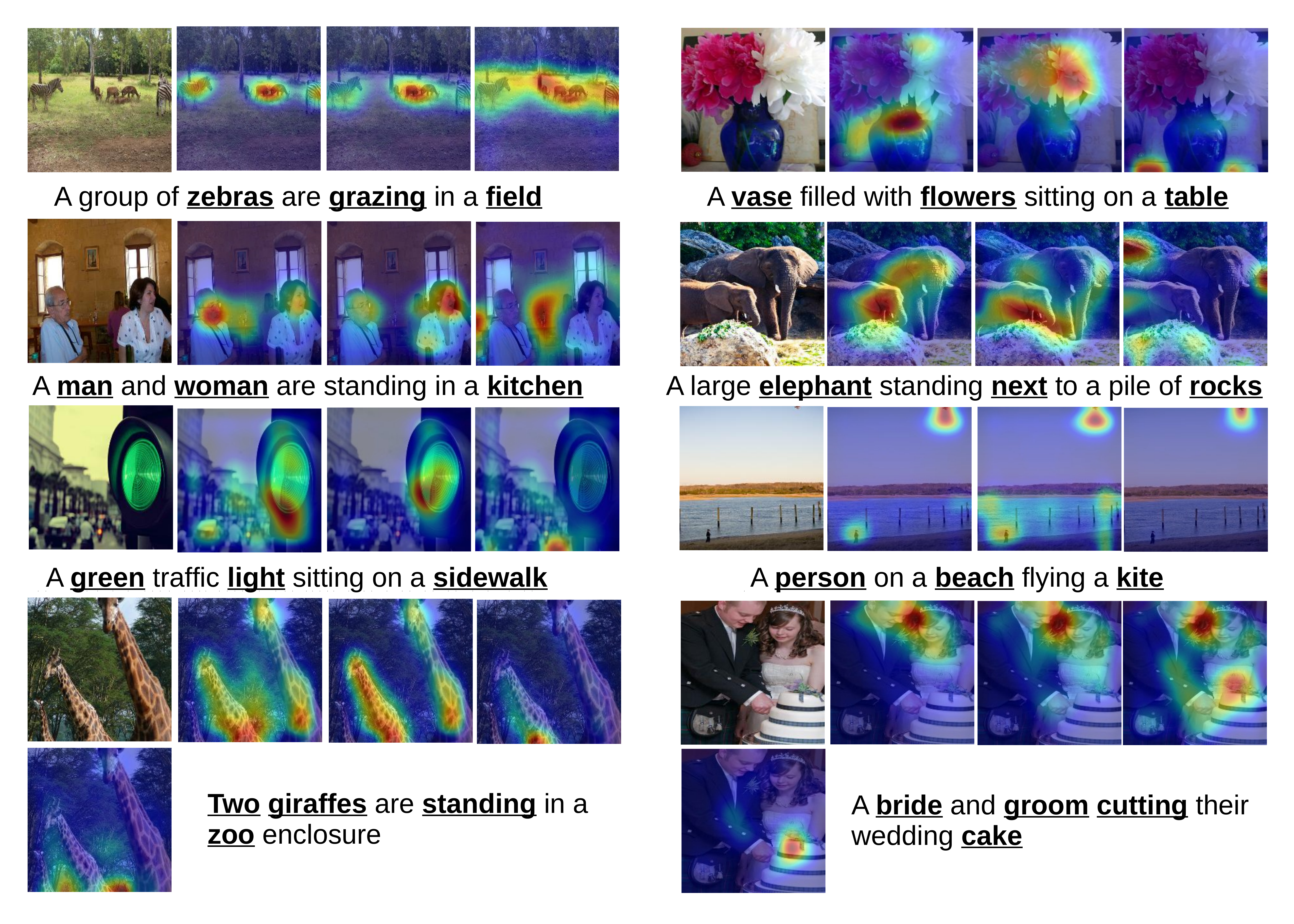}
\vspace{-0.3in}
\caption{Visualization of attention maps learned by our model. % visualization. 
%The attention are learned by our model without hierarchical attention, and 
The left-most images are the input images, and each bold-underline word corresponds to one attention map. \wqz{Here we show nouns, verbs, prepositions and numbers.}}\label{fig6}
\end{figure}

\begin{figure}
\centering
\includegraphics[width=0.5\textwidth]{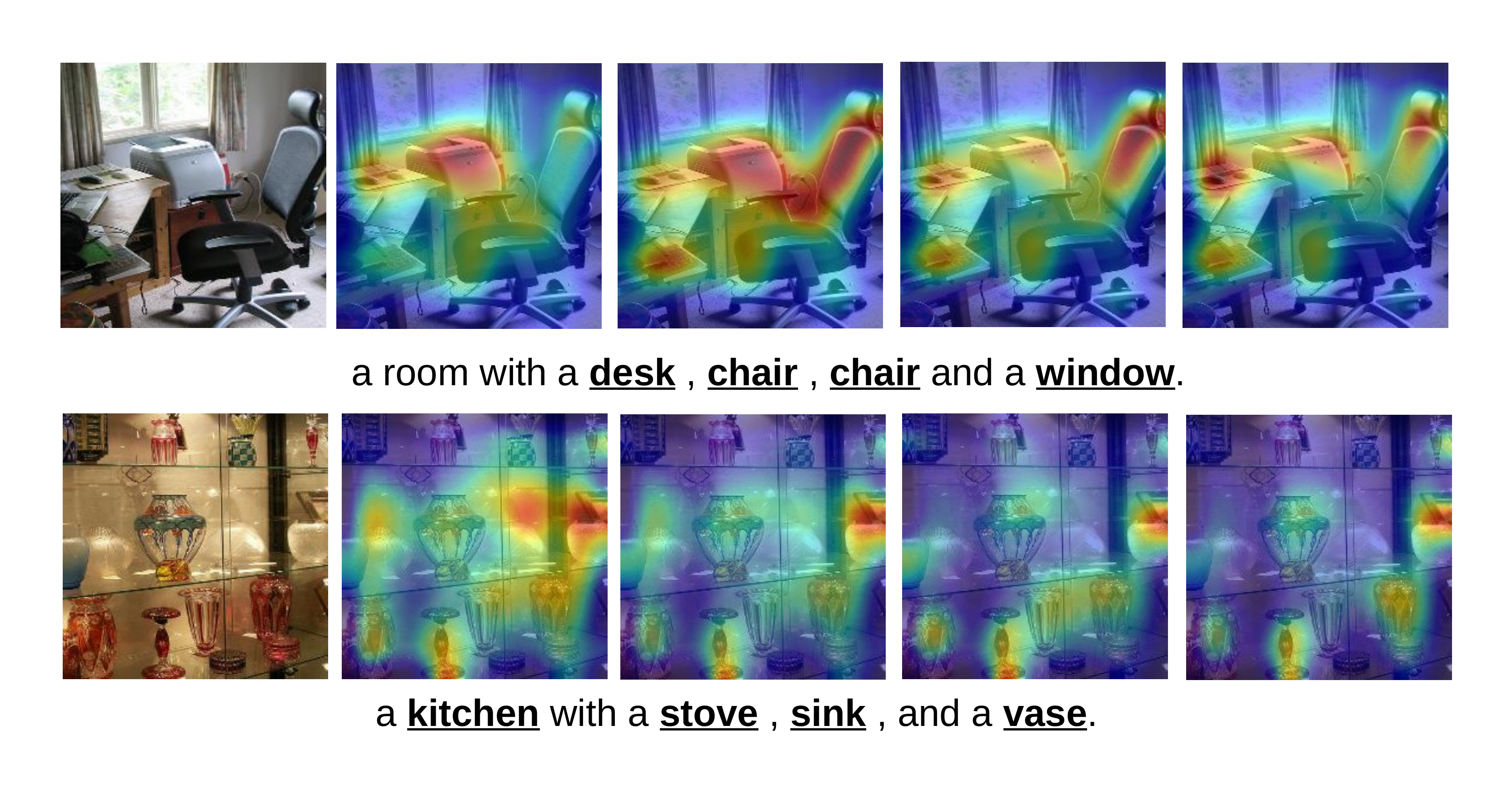}
\vspace{-0.3in}
\caption{Example failure cases. \wqz{Scenes with many objects are challenging for the visual attention model.}}\label{fig7}
\end{figure}

\begin{figure}
\centering
\includegraphics[width=0.4\textwidth]{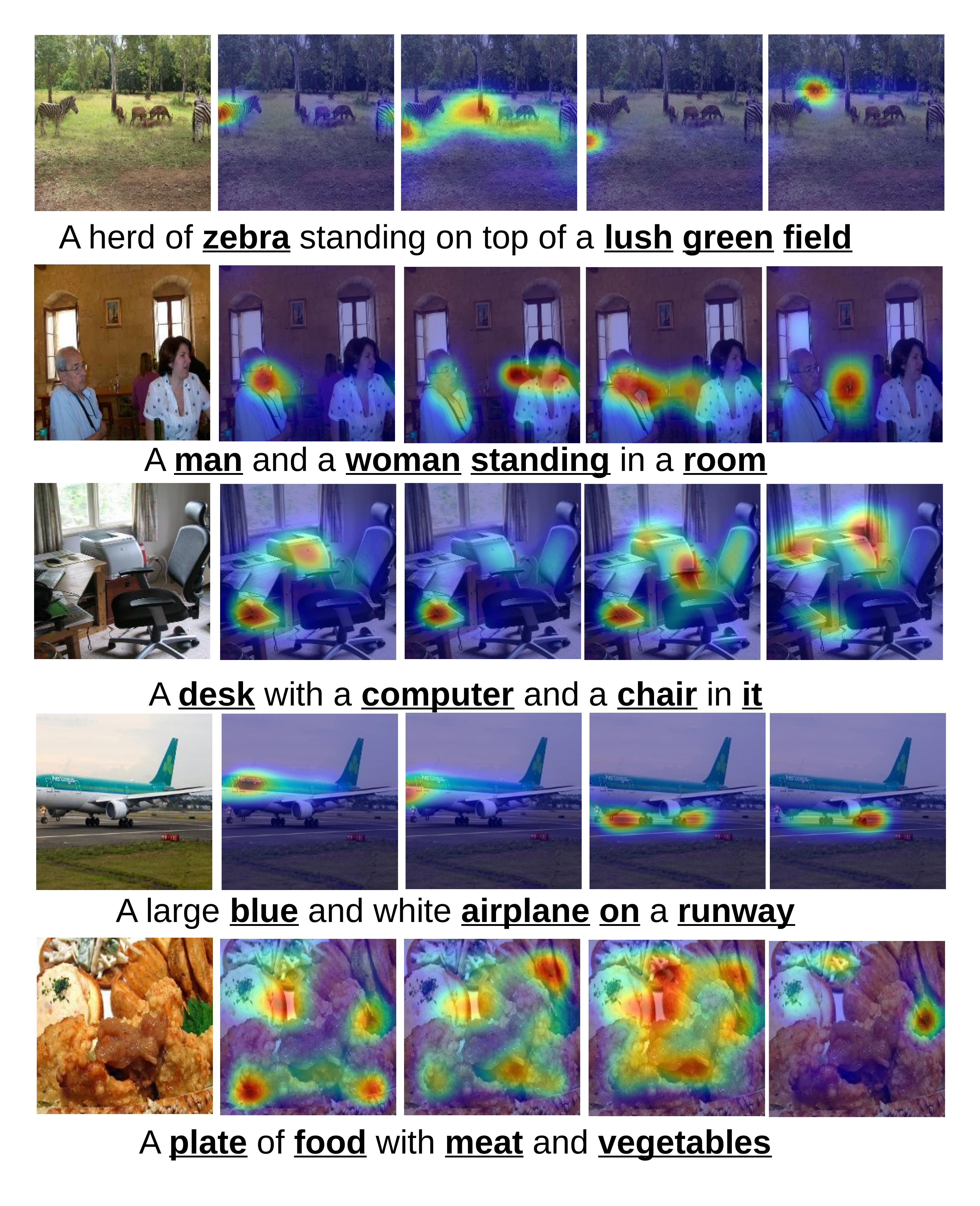}
\vspace{-0.1in}
\caption{%Visualizing the hierarchical attention maps at different level. $rf$ denotes the receptive field size for each layer, which means the nodes at this layer are able to see $rf$ words to the left. The model starts from the starting token $<$S$>$, and the attention generated by layer 6 is used by the prediction module.
\wqz{Visualization of the attention maps at the top-level of our hierarchical attention. The hierarchical attention can handle many-object scenes. %, to some extent.
}}\label{fig8}
\end{figure}

\wqz{We use the MSCOCO test images to visualize the attention maps and descriptions.}
%\NOTE{these examples are from test or training set?}
%
The attention maps are visualized by upsampling the 14$\times$14 attention maps with bilinear interpolation and Gaussian smoothing (see \reffig{fig6}).
%Our attention map is with the size of $14\times 14$, and to visualize we simply upsample the map using bilinear interpolation and Gaussian smooth method. 
%\reffig{fig6} shows the attention map learned by our proposed model without hierarchical attention. From figure \ref{fig6} we can find that
Our proposed model is able to pay attention to correct area when predicting the corresponding words. However, it is difficult to quantify the attention module. \wqz{E.g., when the model predicts the word ``man'' (second row in \reffig{fig6}), it also pays attention to the woman, and vice versa. For the prepositions and verbs, our model attends to the correct area, \eg when the word ``next'' is present, the areas around the elephant are highlighted.}
%which attention is better, %\eg in (a) and (d) when the model predicts the word "zebra", it tends to pay attention to one animal, while in (b) and (c) multiple animal attends and the model predicts the plural form "zebras", and both are reasonable. 
%\NOTE{Why show different values of $k$? Maybe just show the best $k=3$, and show several examples.}

%We also find 
When there are many objects in an image, although the model can pay attention to the areas that contain objects, it predicts the wrong words or repeats a words several times, as seen in \reffig{fig7}. The top row shows that the model predicts the word ``chair'' two times, and the attention maps are similar. The bottom row shows that although the objects are attended, the model predicts the wrong words ``stove'' and ``sink'', which are likely influenced by the preceding context % But if we consider the context 
``a kitchen with a''.

%, the prediction is more likely reasonable.

We also visualize our model with hierarchical attention in \reffig{fig8}, which shows the attention maps at the \wqz{top}-level for \wqz{the predicted words}.
%To figure out whether our hierarchical attention is able to learn different attentions for different concepts learned by each layer of the decoder, we visualize the attention maps at each level. 
%
%In the input layer, the model does not necessarily pay attention to \abc{meaningful} areas. \abc{However, in each successive layer, the attention is refined to cover regions of interest, e.g., the 7th image of layer 1 corresponds to the words ``zebra standing on''.}
% shows that it learns something meaningful.
\wqz{The model with hierarchical attention tends to pay attention to small areas at the top level, and for the scenes with many objects it is able to focus on the relatively correct areas, which is similar to the coarse-to-fine procedure. Compared with the failure cases in \reffig{fig7}, using hierarchical attention provides more details, which could benefit caption generation.}

\vspace{-0.1in}
\section{Conclusions}
\vspace{-0.1in}
We have developed a CNN$+$CNN framework for image captioning and explored the influence of the kernel width and the layer depth of the language CNN. We have shown that the ability of the CNN-based framework is competitive to LSTM-based models, but can be trained faster. We also visualize the learned attention maps to show that the model is able to learn concepts and pay attention to the corresponding areas in the images in a meaningful way.
% and we found that at different level the model is able to learn something meaningful.
%
%Our future work will focus on exploring the ability of CNNs to learn concepts and the relationships between the concepts and images. 
%What's more, we will try to boost the performance of the CNN-based framework and make more comparisons with the LSTM-based models.

%%%%%%%%%%%% references
{\small
\bibliographystyle{ieee}
\bibliography{myreferences}
}

\end{document}